\pdfoutput=1

\documentclass[11pt]{article}

\usepackage[]{acl}

\usepackage{times}
\usepackage{latexsym}
\usepackage{graphicx}
\usepackage{booktabs}
\usepackage{multirow}
\usepackage{subcaption}
\usepackage{caption}
\usepackage{amsfonts}

\usepackage[T1]{fontenc}

\usepackage[utf8]{inputenc}

\usepackage{microtype}

\title{Table-based Fact Verification with Self-adaptive\\ Mixture of Experts}

\author{Yuxuan Zhou$^{1}$, Xien Liu$^{1}$, Kaiyin Zhou$^{2,3}$, Ji Wu$^{1}$ \\
	$^1$Department of Electronic Engineering, Tsinghua University, Beijing 100084, China  \\
	$^2$THiFLY Research, Tsinghua University, Beijing 100084, China  \\
	$^3$State Key Laboratory of Cognitive Intelligence, Hefei, Anhui 230088, China  \\
}

\begin{document}
\maketitle
\begin{abstract}
The table-based fact verification task has recently gained widespread attention and yet remains to be a very challenging problem. It inherently requires informative reasoning over natural language together with different numerical and logical reasoning on tables (e.g., count, superlative, comparative). 
Considering that, we exploit mixture-of-experts and present in this paper a new method: \textbf{S}elf-\textbf{a}daptive \textbf{M}ixture-\textbf{o}f-\textbf{E}xperts Network (\textbf{SaMoE}).
Specifically, we have developed a mixture-of-experts neural network to recognize and execute different types of reasoning---the network is composed of multiple experts, each handling a specific part of the semantics for reasoning, whereas a management module is applied to decide the contribution of each expert network to the verification result. A self-adaptive method is developed to teach the management module combining results of different experts more efficiently without external knowledge. The experimental results illustrate that our framework achieves \textbf{85.1\%} accuracy on the benchmark dataset \textsc{TabFact}, comparable with the previous state-of-the-art models. We hope our framework can serve as a new baseline for table-based verification. Our code is available at \url{https://github.com/THUMLP/SaMoE}.
\end{abstract}
\section{Introduction}
\begin{figure}[htbp]
	\centering
	\includegraphics[scale=0.57]{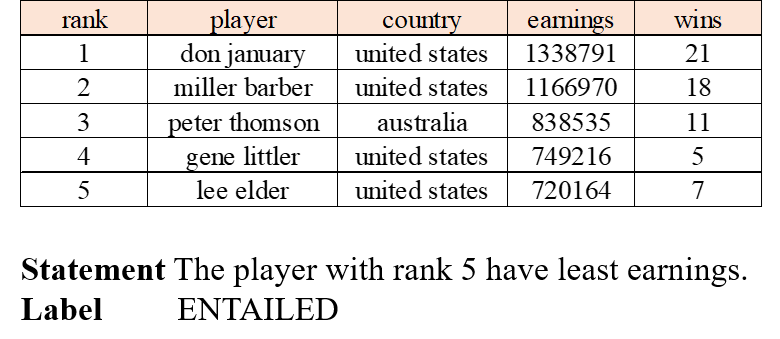}
	\caption{An Example of table-based fact verification.}
	\label{fig:verification example}
\end{figure}

Fact Verification, aiming to determine the consistency between a statement and given evidence, has become a crucial part of various applications such as fake news detection, rumor detection \citep{rashkin-etal-2017-truth,Thorne18Fever,goodrich2019assessing,vaibhav-etal-2019-sentence, kryscinski-etal-2020-evaluating}. While most existing research focuses on verification based on unstructured text, a new trend is extending the scope to structured evidence (e.g., tables), which is informative and ubiquitous in our daily lives. 
Table-based verification faces different challenges than unstructured-text-based due to the complexity of the requirements, including sophisticated textual, numerical, and logical reasoning across evidence tables; even for some statements, multiple types of reasoning are indispensable to complete the verification. An example is presented in Figure \ref{fig:verification example}.

To tackle the challenges above, previous work established two kinds of methods: (1) \textit{program-enhanced methods} \citep{2019TabFactA, zhong-etal-2020-logicalfactchecker, shi-etal-2020-learn,yang-etal-2020-program} and (2) \textit{table-based pre-trained models} \citep{eisenschlos-etal-2020-understanding, liu2021tapex}.
The program-enhanced methods mainly leverage programs generated by the semantic parser. Specifically, statements are parsed into executable programs to extract the logical/numerical semantics, which is further be leveraged together with contextual semantics learned by a language model (e.g., BERT) in inference. However, the semantic parsers that generate semantic-consistent programs must be trained in a weak supervision setting, which brings difficulties in training. Furthermore, generalizing this method to other datasets is almost impossible without the API set modification according to the reasoning requirements on the new datasets. 

\begin{figure*}[t]
	\centering
	\includegraphics[scale=0.5]{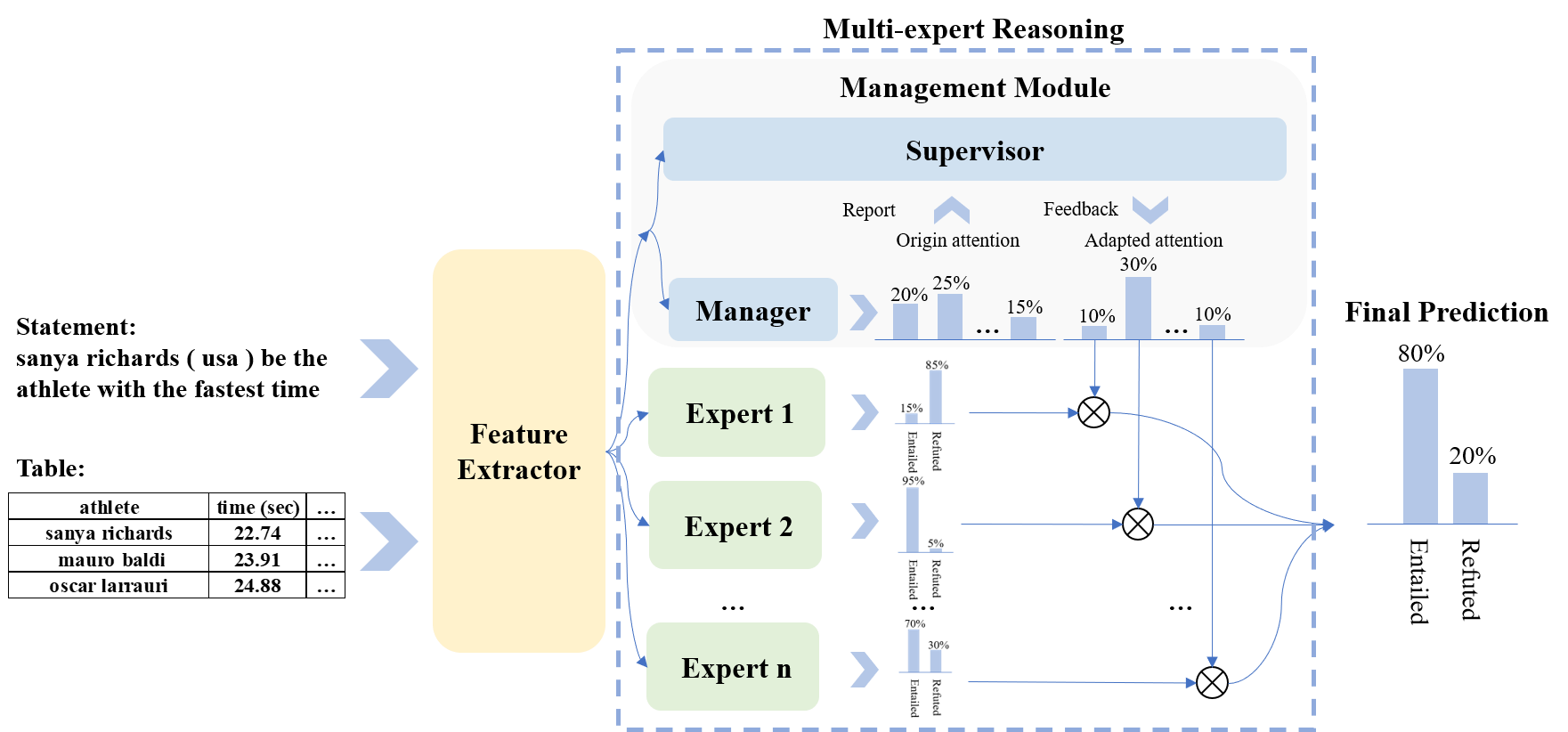}
	\caption{An overview of \textbf{S}elf-\textbf{a}daptive \textbf{M}ixture-\textbf{o}f-\textbf{E}xperts Network (\textbf{SaMoE}) for table-based fact verification.}
	\label{fig:overview}
\end{figure*}
The table-based pre-trained models leverage elaborate model structure \citep{herzig2020tapas} and pre-training tasks \citep{eisenschlos-etal-2020-understanding, liu2021tapex} to enhance the reasoning skills on structured data. Nevertheless, two significant shortcomings remain. Firstly, the process is demanding due to the tremendous computing resources required by pre-training. Moreover, the effectiveness of pre-training to its downstream tasks mainly depends on the adaptability between these two tasks. Therefore, implementing pre-training tasks may fail to meet the requirements when facing the unseen reasoning types demanded by new datasets.

In this paper, we introduce an innovative framework, \textbf{S}elf-\textbf{a}daptive \textbf{M}ixture \textbf{o}f \textbf{E}xperts (SaMoE), to address the previously mentioned problems. 
The entire framework is illustrated in Figure \ref{fig:overview}.
SaMoE consists of 3 components: \textbf{feature extractor}, \textbf{experts}, and \textbf{management module}, which is the combination of \textbf{manager} and \textbf{supervisor} networks. Each expert initially takes the same feature as input and then learns to deal with different parts of the reasoning types (e.g., contextual/logical/numerical) required by table-based verification.
A management module is designed to guide the training of experts and combine experts' verification results effectively. 
The manager network in this module assigns each expert a unique attention score, allowing each individual to focus on different kinds of reasoning types and summarizes experts' entire outputs as the final verification result.
However, managers failed to allocate the highest attention score to the expert who performs best on the current reasoning type in most circumstances. 
To alleviate this problem, we introduce a supervisor network to adjust the attention score given by the manager.
The supervisor network is trained \textbf{self-adaptively} (i.e., it learns directly from experts' performance on the train set) without prior knowledge of the task or dataset. 
Extensive experiments are conducted to show that our proposed framework, implemented with a general pre-trained language model RoBERTa \citep{liu2019roberta}, outperforms previous state-of-the-art methods, including table-based pre-trained models. The main contributions of this work are as follows:

\begin{itemize}
    \item We innovatively implement mixture-of-experts for table-based verification, aiming to arrange each expert to different types of reasoning. This method can also be easily generalized to other datasets. 
    \item We investigate a self-adaptive method to adjust suitable attention score to each expert according to its performance on different reasoning types, achieving more efficient cooperation across experts.
    \item Our framework achieves better performance on the \textsc{TabFact} dataset without the assistance of table-based pre-trained models.
\end{itemize}
\section{Research Question}
The table-based verification task expects one to determine whether a statement $S$ is entailed or refuted by an evidence table $T$. The process above can be regarded as a binary classification task and thus denoted as $f(S, T)=\hat{y}$, where $f$ is the verification model and $\hat{y}\in\{0,1\}$ its prediction.
\section{Methods}
We present the proposed framework (SaMoE), which leverages a set of experts to deal with different parts of the reasoning types involved in table-based verification. This section is organized as follows. Sec.\ref{sec: feature} introduces the feature extractor that extracts the joint semantics of the table-statement pair. Sec.\ref{sec: experts} describes experts that verify the statements separately based on the same extracted semantics. Sec.\ref{sec: management} describes the management module that guides the experts' training and combines their verification results effectively; two components of this module, the manager and the supervisor, are introduced in this section individually.
\subsection{Feature Extractor} \label{sec: feature}
Feature extractor parses the statement-table pair and learns the joint table-statement semantics. Tables are initially pruned and serialized into a sequence. Subsequently, the serialized tables are transmitted into the language model together with the statements for joint representation learning. These two processes will be further interpreted in the following subsections. 
\subsubsection{Table Pre-processing}
As for Tables, the pre-processing (pruning and serializing) before the joint representation learning provides convenience for subsequent processing of the existing language model.
\paragraph{Table Pruning}
Table pruning discards some parts of the table that do not participate in the verification, according to the input size limit of the language model. 
We take advantage of the table-pruning algorithm proposed in \citet{2019TabFactA} and further enhance its performance.
The original algorithm matches the entities in statements with cells in tables by a heuristic method and selects the columns that include matched cells to form the pruned table. 
Noticed that the algorithm always fails to select the critical columns of verification while there is still room left for the input sequence of the language model, we further add a greedy strategy on the algorithm that keeps adding columns that are not selected to the pruned table until reaching the maximum input size of the downstream model to make the best use of its capacity.
\paragraph{Table Serializing}
Tables are further serialized to a 1-D sequence after pruning to be compatible with the input format of the language model. We follow the serializing method used in TABLE-BERT \citep{2019TabFactA} that paraphrases tables with a natural language template.
Specifically, a table with $m$ rows and $n$ columns is paraphrased as ``row 1 is: $h_1$ is $T_{11}$; ... ; $h_n$ is $T_{1n}$. row 2 is: ... row $m$ is: $h_1$ is $T_{m1}$; ... ; $h_n$ is $T_{mn}$.", where $h_i$ refers the $i^{th}$ header and $T_{ij}$ the value in the $(i, j)-th$ cell of table $T$. 
We find that such template-serialized tables are more suitable for language models pre-trained on unstructured text to process.
\subsubsection{Joint Representation Learning}
After the table pre-processing, the serialized table and the statement are further passed to a language model to learn the joint contextual representation of each token.
The learned representation vectors are then transmitted to the experts and the management module for inference and management. Specifically, the serialized table and the statement are initially tokenized into two token sequences \textbf{\~T} and $\textbf{S}$. Then the joint token sequence $\textbf{X}$ is formed as $\textbf{X}=[\langle s\rangle$, $\textbf{S}$, $\langle /s\rangle$, \textbf{\~T}, $\langle /s\rangle]$, where $\langle s\rangle$ and $\langle /s\rangle$ are the separators that identify the beginning and the end of each token sequence. 
The token sequence $\textbf{X}$ will be padded or truncated to fit the maximum input length of the language model. Finally, a transformer model is applied to learn the contextual representation of $\textbf{X}$ :
\begin{equation}
    \textbf{H}=f_{LM}(\textbf{X})  
\end{equation}
where $\textbf{H} \in \mathbb{R}^{n\times d}$ refers to the learned joint representation, $n$ is the maximum input length and $d$ the dimension of the representation vector. $f_{LM}$  denotes the contextual representation learning process of the language model. 
In this paper, we implement it with \textit{transformer} \citep{vaswani2017attention}, the most popular contextual representation model in recent years. 
\subsection{Experts}\label{sec: experts}
A group of experts is applied to verify the statements separately based on the same statement-table joint semantics extracted by the feature extractor module. Experts share the same model structure, while the parameter learning strategy of SaMoE gives expert differentiation. 
Specifically, each expert is implemented with a stack of transformer encoding layers. An MLP classifier that calculates the probability of the statement is entailed by the evidence table based on the encoded semantics.
We implement experts with the same general structure rather than different structures specially designed for certain reasoning types since we anticipate that the proposed framework can be smoothly generalized to other datasets.
The process above can be formulated as follows:
\begin{equation}
    \textbf{h}_{i}=f_{Enc_i}(\textbf{H})
\end{equation}
\begin{equation}
    \textbf{p}_{i}=softmax(tanh(\textbf{h}_{i}\textbf{W}_{1}^{i})\textbf{W}_{2}^{i})
\end{equation}
where $\textbf{h}_{i}\in \mathbb{R}^{d}$ is the token $\langle s\rangle$'s final representation vector encoded by the $i^{th}$ expert's encoder $Enc_i$. It implies the $i^{th}$ expert's whole understanding to the statement-table pair. $\textbf{W}_{1}^{i} \in \mathbb{R}^{d \times d}$ and $\textbf{W}_{2}^{i} \in \mathbb{R}^{d \times 2}$ are the trainable parameters of $i^{th}$ expert's classifier, which projects $\textbf{h}_{i}$ to the probabilities $\textbf{p}_{i} \in \mathbb{R}^{2}$ that the statement is entailed/refuted by the table. $tanh$ and $softmax$ are activation functions. $n_e$ refers to the number of experts.
\subsection{Management Module} \label{sec: management}
Learning the joint semantics parsed in Sec.\ref{sec: feature}, the management module intends to generate attention scores to bias experts' training and combine experts' results efficiently. The module consists of two components: manager and supervisor, both of them are implemented based on \textit{transformer} model. The manager is mainly designed to guide experts' training, while the supervisor is applied to combine experts' results efficiently.
\paragraph{Manager} \label{sec: manager}
The manager guides the training of experts and forms a preliminary assumption to the expert combination. It encodes the joint representation matrix and generates attention scores $\textbf{a}_{M}$ to guide the experts' training process:
\begin{equation}
    \textbf{h}_{M}=f_{Enc_M}(\textbf{H})
\end{equation}
\begin{equation}
    \textbf{e}_{M}=tanh(\textbf{h}_{M}\textbf{W}_{1}^{M})\textbf{W}_{2}^{M}
\end{equation}
\begin{equation}
    \textbf{a}_{M}=softmax(\textbf{e}_{M})
\end{equation}
where $Enc_M$ denotes the manager's encoder, $\textbf{W}_{1}^{M} \in \mathbb{R}^{d\times d}$ and $\textbf{W}_{2}^{M} \in \mathbb{R}^{d\times n_e}$ are trainable parameters. The network structures of the manager and experts are basically the same, only different in the layers of the encoder and the output dimension.

After preceding calculation, the normalized attention scores $\textbf{a}_{M}$ are used to guide the training of experts by a specially designed verification loss, which will be introduced in Sec.\ref{sec: verification loss}.
\paragraph{Supervisor} \label{sec: supervisor}
The supervisor adjusts the attention scores submitted by the manager to improve the cooperative efficiency among experts (i.e. assigning higher weights to experts who perform better on the current input pair). %
The network predicts the deviation between the preliminary assumption (i.e., the attention) and the ideal combination weights based on the knowledge encoded in the joint representation matrix $\textbf{H}$:
\begin{equation}
    \textbf{h}_{S}=f_{Enc_S}(\textbf{H})
\end{equation}
\begin{equation}
    \textbf{e}_{S}=tanh(\textbf{h}_{S}\textbf{W}_{1}^{S})\textbf{W}_{2}^{S}
\end{equation}
\begin{equation}
    \textbf{a}_{S}=softmax(\textbf{e}_{M}+\textbf{e}_{S})
\end{equation}
where $\textbf{W}_{1}^{S} \in \mathbb{R}^{d\times d}$, $\textbf{W}_{2}^{S} \in \mathbb{R}^{d\times n_e}$ are trainable parameters and $Enc_S$ refers to the encoder of the supervisor. Parameters of the supervisor are optimized self-adaptively based on experts' performance on the train set. More details of this learning strategy will be presented in Sec.\ref{sec: self-adaptive}.
\section{Parameter Learning}
Parameters in SaMoE are learned in two consecutive stages: 1) Multi-expert training: parameters in the feature extractor, experts and the manager are end-to-end optimized under the supervision of labels; 2) Self-adaptive learning: parameters in the supervisor are self-optimized by observing experts' performance on the train set (other parameters are fixed simultaneously). A weighted sum of two losses is minimized in the first stage to achieve diverse and balanced training of experts. For the second stage, we minimize a self-adaptive loss calculated based on the experts' classification loss. Subsequent sections introduce these two learning stages in detail.
\subsection{Multi-expert Training}\label{sec:supervised learning}
Multi-expert training guides each expert on dealing with different reasoning types while maintaining balanced training across experts. 
To achieve the goals above, we develop two loss functions: 1) verification loss $\mathcal{L}_V$ that measures the weighted sum of each expert's classification loss, differentiating experts' learning with different attention scores assigned by the manager; 2) manager assumption loss $\mathcal{L}_M$ that is applied to prevent the occurrence of imbalanced training across experts.
The overall loss of this state is calculated by a weighted sum of these two terms: $\mathcal{L}_1=\mathcal{L}_V+\lambda \mathcal{L}_M$, where $\lambda$ is a hyperparameter that controls the ratio of $\mathcal{L}_M$. Subsequent sections provide detailed introduction to these two terms.
\subsubsection{Verification Loss}\label{sec: verification loss}
The verification loss $\mathcal{L}_V$ is designed based on the loss function proposed in \citet{jacobs1991adaptive}. It is calculated by the weighted sum of each expert's cross-entropy:
\begin{equation}
    \mathcal{L}_V=\sum_{i=1}^{n_e}(\textbf{a}_M)_{i}\cdot CE(\textbf{p}_{i},l)
\end{equation}
where $(\textbf{a}_M)_{i}$ is the $i^{th}$ element of the attention scores $\textbf{a}_M$, $l\in\{0,1\}$ is the label of the statement-table pair and $CE(\cdot,\cdot)$ the cross-entropy function. Note that it is necessary to calculate each expert's cross-entropy independently. We want each expert to behave like an independent expert (i.e., complete the verification without the help of other experts). The attention vector $\textbf{a}_M$ acts as a "training scheduler" in this loss function: experts that are assigned with larger attention scores receive a larger gradient than other experts on the current input, resulting in diverse experts' performance.
\subsubsection{Manager Assumption Loss}\label{sec:manager assumption}
\begin{figure}[htb]
	\centering
	\includegraphics[scale=0.28]{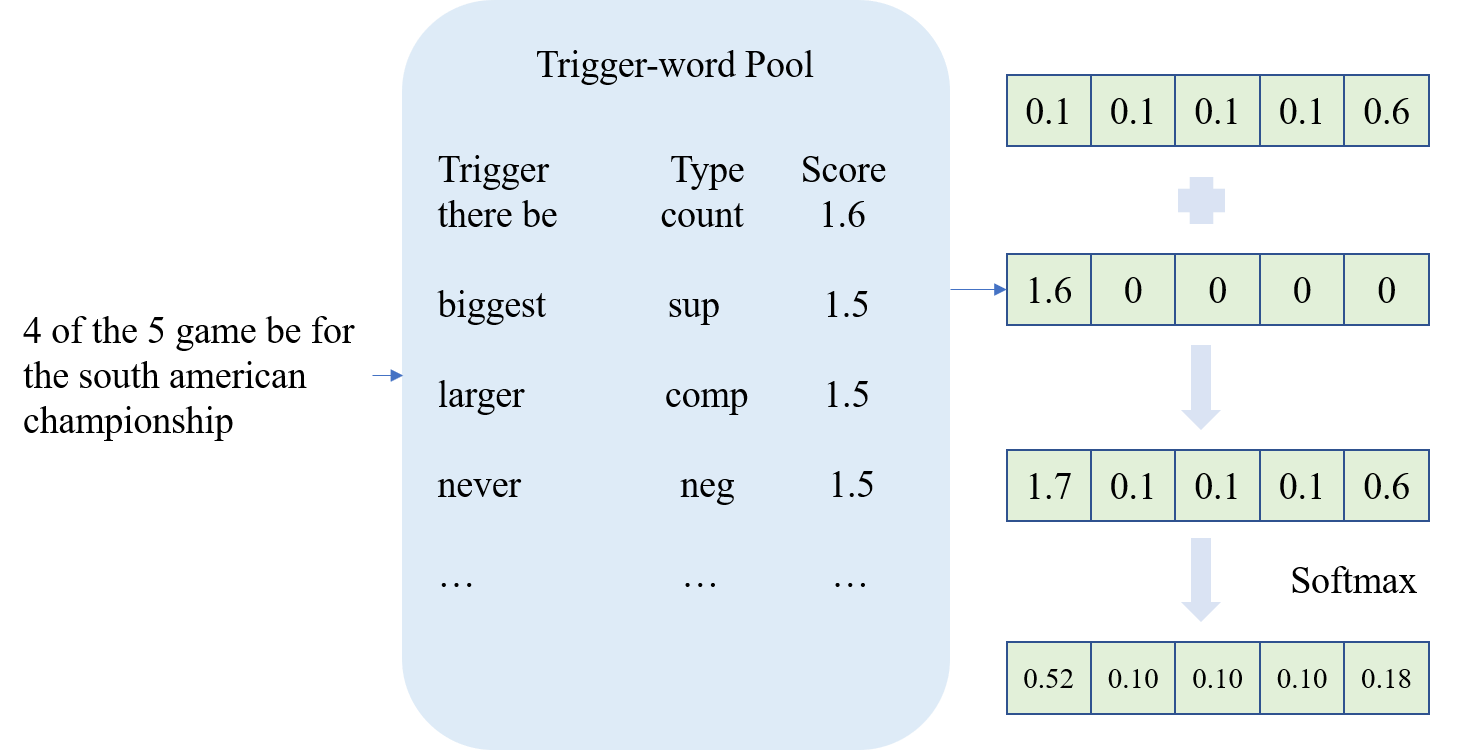}
	\caption{An example of prior assumption generation with $n_e=5$ and 4 predefined reasoning types.}
	\label{fig:prior}
\end{figure}

We have trained the MoE with only the verification loss $\mathcal{L}_V$ and observe a severe "imbalanced experts" phenomenon that only one expert is well-trained (i.e., the expert performance is improved by training) and the manager keeps assigning a close-to-1 attention score to this expert, which is also reported in previous research \citep{eigen2013learning, shazeer2017outrageously}. To avoid this problem, we develop another loss function that forces the manager to assign reasonable attention scores to experts:
\begin{equation}
    \mathcal{L}_{M} = D(\textbf{a}_{P}||\textbf{a}_M)
\end{equation}
where $D(\cdot||\cdot)$ denotes the Kullback–Leibler divergence and $\textbf{a}_{P}$ a prior assumption that is generated with a simple heuristic algorithm (to be introduced in the next paragraph) which requires limited prior knowledge of the reasoning types. By minimizing $\mathcal{L}_{M}$, the manager learns to assign each expert with a reasonable attention score, resulting in a balanced training across experts.
\paragraph{Prior Assumption Generation}
The prior assumption $\textbf{a}_{P}$ is generated to represent the probabilities that the statement involves different reasoning types that we are interested in.  Specifically, we develop a trigger-word-based heuristic algorithm to form the prior assumption for each statement automatically:
\begin{enumerate}
    \item Initialize the prior assumption with $\textbf{e}_{0}\in \mathbb{R}^{n_e}$, which is empirically set as $(0.1, 0.1, ... ,0.6)^T$. The $(\textbf{e}_{0})_{n_e}$ represents the probability that the statement does not involve any predefined reasoning types and thus is set higher than other values in advance.
    \item Traverse the trigger-word set\footnote{a set of words that suggest a specific reasoning type, see Appendix \ref{apdix:specific prior} for more information.} of each reasoning type ($n_e-1$ types in total). If a trigger word/pattern $w$ that belongs to  $i^{th}$ reasoning type is detected in the statement, the trigger's weight $s_w$ (set empirically) is accumulated to the $i^{th}$ dimension of a zero-initialized bias vector $\delta \in \mathbb{R}^{n_e}$: $\delta_i \gets \delta_i+s_w$.
    \item Add the bias vector $\delta$ to the prior assumption $\textbf{e}_{0}$ and normalize to get the prior assumption: $\textbf{a}_{P}=softmax(\textbf{e}_{0}+\delta)$.
\end{enumerate}
Figure \ref{fig:prior} presents an example of this process. Learning to imitate the prior assumption, the manager guides each expert to focus on different reasoning types and thus achieves diverse experts. We implement a relatively small trigger-word pool in experiments and find the method works effectively, indicating that the method can be smoothly generalized to other datasets with little modification to the predefined reasoning types and trigger-word pool.
\subsection{Self-adaptive Learning}\label{sec: self-adaptive}
Self-adaptive learning aims to enhance further the expert combining efficiency with only the knowledge of the expert's performance on the train set. Specifically, an ``expert ability" vector $\textbf{a}_E \in \mathbb{R}^{n_e}$ is calculated based on the ``expert loss" vector $\textbf{m} \in \mathbb{R}^{n_e}$, where $\textbf{m}_i=CE(\textbf{p}_{i},l)$ is the cross-entropy loss of the $i^{th}$ expert. Note that the cross-entropy of the expert is negatively correlated with its performance. Then the expert ability vector $\textbf{a}_E$ is calculated as follows:
\begin{equation}
    \textbf{a}_E = softmax(-\alpha\cdot\textbf{m})
\end{equation}
where $\alpha=\sqrt{\beta/Var(\textbf{m}})$ is a variance-normalizing coefficient and $\beta$ is a hyperparameter that decides the variance of the expert ability vector before the activation (i.e., $Var(-\alpha\cdot\textbf{m})=\beta$). Such normalization is designed based on the observation that $\textbf{m}$ tends to have a extreme small variance and $softmax(-\textbf{m})$ often generates a close-to-uniform distribution. Note that the generated $\textbf{a}_E$ is positively correlated with the experts' performance (e.g., if the $i^{th}$ expert outperforms the $j^{th}$ expert on the input pair then we have $(\textbf{a}_E)_i > (\textbf{a}_E)_j$).

Based on $\textbf{a}_E$, we develop the loss function that has the same form with $\mathcal{L}_M$ in Sec \ref{sec:manager assumption}:
\begin{equation}
    \mathcal{L}_{S}=D(\textbf{a}_E||\textbf{a}_S)
\end{equation}

By minimizing the loss above, the higher attention scores are assigned to the best-performed experts after the supervisor's adjustment, resulting in more efficient cooperation across experts.
\section{Experiment Setup} \label{sec: experiments}
\subsection{Data and Metric}
We conduct the experiments on \textsc{TabFact}, a large scale benchmark dataset of the table-based fact verification task\footnote{We did not conduct experiments on other datasets such as SEM-TAB-FACTS \citep{wang-etal-2021-semeval} and InfoTabs \citep{gupta-etal-2020-infotabs}, since there is little work and comparisons have been made on these datasets.}. \textsc{TabFact} contains a total of 117k statements and 16k Wikipedia tables. The test set is further divided into a simple and complex subset based on verification difficulty, for verifying some statements on \textsc{TabFact} requires more logical/numerical reasoning skills. We choose accuracy as the evaluation metric following the existing work to make our experiment results comparable. More details of \textsc{TabFact} are presented in Appendix \ref{apdix:statistics}. 
\subsection{Implementation Details}
\paragraph{Training Details}
We set $n_e=5$ expert networks in our implementation of SaMoE. The transformer layers are 12 for encoders in the feature extractor and experts and 2 for encoders in the manager and supervisor. The hidden states' dimension $d$, the maximum input length $n$, the $\lambda$ in Sec.\ref{sec:supervised learning}, and the $\beta$ in Sec.\ref{sec: self-adaptive} are set to 1024, 512, 0.1 and 0.1 respectively. We applied RoBERTa-Large \citep{liu2019roberta} to initialize the feature extractor and experts in our framework. The details of parameter initialization can be found in Appendix \ref{apdix:parameter init}.

We apply Adam optimizer \citep{kingma2014adam} in training with learning rate 2e-5, dropout rate 0.1, warmup step 17,304, and batch size 32. SaMoE is first trained in the Multi-expert training stage for 57,680 steps (20 epochs). Then the supervisor is trained in the self-adaptive learning stage for another 5,000 steps, while the best parameters of other parts in the framework are loaded and fixed. 
The model is evaluated every 1000 steps, and the model that achieves the highest performance on the development set is saved.
All the codes are implemented with Pytorch \citep{paszke2019pytorch} and the transformers package \citep{wolf-etal-2020-transformers}. 
We train all our models on a single GeForce RTX 3090.
\paragraph{Settings of Prior Assumption}
We choose the top 4 types of reasoning types that appear most frequently in \textsc{TabFact}\footnote{We follow the statistics in \citet{2019TabFactA} for the frequency of different reasoning types.} (count, comparative, superlative, negation). We apply a small trigger-word pool containing only 26 trigger words, injecting limited prior knowledge of the dataset. More details of this part are presented in Appendix \ref{apdix:specific prior}.
\subsection{Baselines}
We compared our proposed framework with different kinds of baselines on \textsc{TabFact}: (1) Program-enhanced methods: LPA \citep{2019TabFactA}, LogicalFactChecker \citep{zhong-etal-2020-logicalfactchecker}, HeterTFV \citep{shi-etal-2020-learn}, ProgVGAT \citep{yang-etal-2020-program} and Decomp \citep{yang2021exploring}; (2) Table-based pre-trained models: \textsc{TaPas} \citep{eisenschlos-etal-2020-understanding} and \textsc{TaPEx} \citep{liu2021tapex}; (3) Other methods: Table-BERT \citep{2019TabFactA} and SAT \citep{zhang-etal-2020-table}. 
\begin{table*}[t]
\centering
\setlength{\tabcolsep}{4mm}{
	\begin{tabular}{lccccc}
		\hline
		Model&Val&Test&Test$_{simple}$&Test$_{complex}$&Small Test\\
		\hline
		TABLE-BERT&66.1&65.1&79.1&58.2&68.1\\
		LPA&65.1&65.3&78.7&58.5&68.9\\
		LogicalFactChecker&71.8&71.7&85.4&65.1&74.3\\
		HeterTFV&72.5&72.3&85.9&65.7&74.2 \\
		SAT&73.3&73.2&85.5&67.2&-\\
		ProgVGAT&74.9&74.4&88.3&67.6&76.2\\
		Decomp-\textsc{Large}&82.7&82.7&93.6&77.4&84.7\\
		\hline
		\textsc{TaPas}-\textsc{Large}&81.5&81.2&93.0&75.5&84.1\\
		\textsc{TaPEx}&\textbf{84.6}&84.2&\textbf{93.9}&79.6&85.9\\
		\hline
		\textbf{SaMoE}&84.2&\textbf{85.1}&93.6&\textbf{80.9}&\textbf{86.7}\\
		\hline
		Human Performance&-&-&-&-&92.1\\
		\hline
\end{tabular}}
\caption{Comparative performance (accuracy) on \textsc{TabFact}.}
	\label{table:Overall performance}
\end{table*}
\section{Results}
\subsection{Overall Performance}
We compare the proposed SaMoE with different kinds of baselines, and the results are listed in Table \ref{table:Overall performance}. Baselines are presented with the best performance reported in the corresponding papers. SaMoE obtains an accuracy of \textbf{85.1\%} on the test set, achieving a new state-of-the-art on the dataset. Results show that our method consistently outperforms all the program-enhanced methods with a significant \textbf{2.4\%} improvement compared with the Decomp method (the best performed program-enhanced method). Note that SaMoE performs similar with Decomp-\textsc{Large} on the simple subset of the test set (93.6\% vs. 93.6\%) while outperforms Decomp-\textsc{Large} with a remarkable \textbf{3.5\%} on the complex subset (80.9\% vs. 77.4\%). Such analysis indicates that the performance improvement is mainly derived from successfully verifying complex statements, which required more sophisticated reasoning than statements in the simple set. SaMoE even shows comparable performance with the previous SOTA \textsc{TaPEx} that is pre-trained to execute SQL queries on tables. Our method outperforms \textsc{TaPEx} with a 0.9\% improvement on the test set and a further 1.3\% improvement on the complex subset, indicating that SaMoE, based on a \textbf{text-based} pre-trained model, performs even better than \textbf{table-based} pre-trained models on a variety of complex reasoning types demanded by the table-based verification. 
\begin{table}[htb]
\centering
\setlength{\tabcolsep}{4mm}{
	\begin{tabular}{lcc}
		\hline
		Model&Val&Test\\
		\hline
		SaMoE&\textbf{84.2}&\textbf{85.1}\\
		SaMoE w/o Sa & 84.0 & 84.7\\
		SaMoE w/o Sa ($n_e=1$) & 83.6 & 84.0\\
		\hline
\end{tabular}}
\caption{Ablation results that shows the effectiveness of the proposed MoE and self-adaptive learning methods. SaMoE w/o Sa denotes that the framework without self-adaptive learning, and $n_e=1$ denotes that the framework involves only one expert, where the management module does not work in this situation.}
	\label{table:Ablation study}
\end{table}
\subsection{Ablation Study}
We further investigate the effectiveness of the MoE structure and self-adaptive learning with an ablation study. We conduct two experiments: one reduces the number of experts to 1 to disable the contribution from the MoE structure (SaMoE w/o Sa ($n_e=1$)); the other trains the proposed framework with only the Multi-expert training stage (SaMoE w/o Sa). Results are presented in Table \ref{table:Ablation study}. The MoE structure achieves a 0.7\% improvement on the test set (84.7\% vs. 84.0\%), and self-adaptive learning further improves the performance slightly (85.1\% vs. 84.7\%). Note that the slight improvement of self-adaptive learning is expected since the experts and the feature extractor are fixed in this stage. The results demonstrate the effectiveness of both the MoE structure and the self-adaptive learning.
\subsection{Effectiveness Analysis}
We show in this section that the effectiveness of the proposed framework is derived from two aspects: the \textbf{differentiation of experts} (each expert outperforms others on a specific part of reasoning types) and the \textbf{effective attention assignment} by the management module (the best-performed experts are assigned with higher attention scores). 
\begin{figure}[htbp]
    \centering
    \begin{subfigure}[b]{0.23\textwidth}
    \centering
    \includegraphics[width=0.9\textwidth]{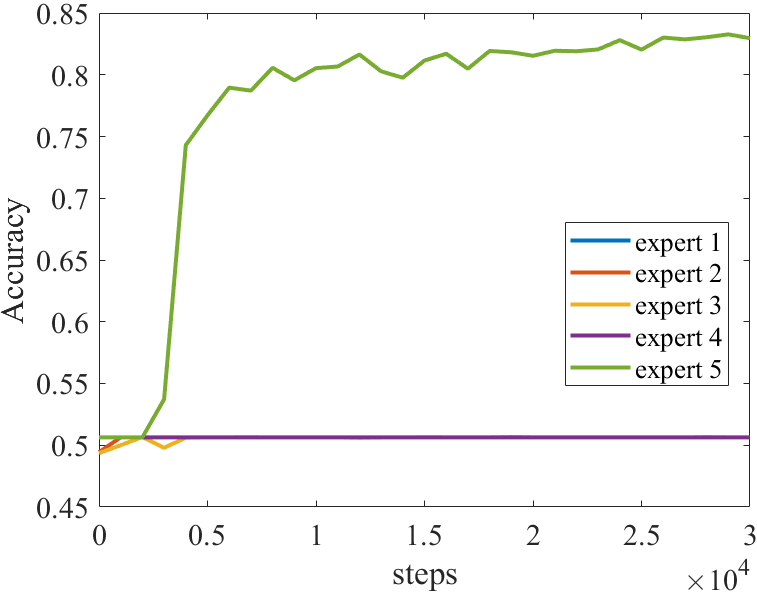}
    \caption{Trained with $\mathcal{L}_V$}
    \end{subfigure}
    \begin{subfigure}[b]{0.23\textwidth}
    \centering
    \includegraphics[width=0.9\textwidth]{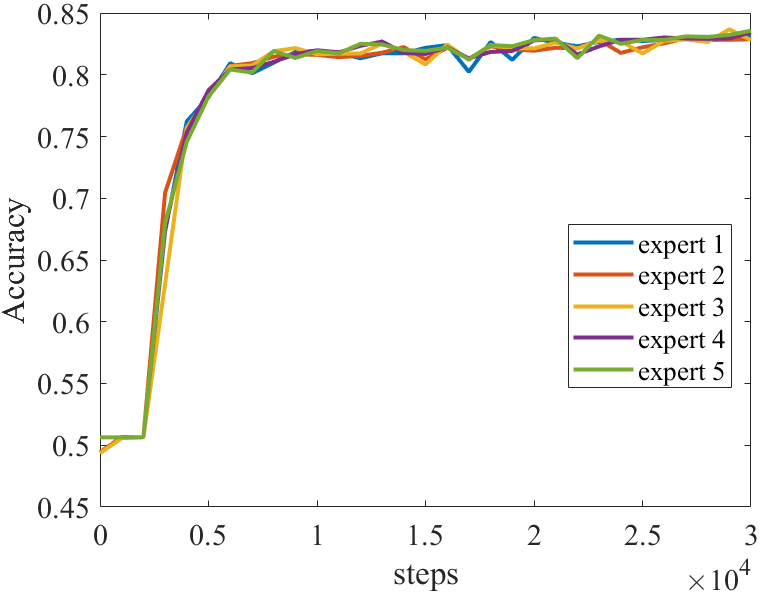}
    \caption{Trained with $\mathcal{L}_V+\mathcal{L}_M$}
    \end{subfigure}
    \caption{Comparison of models trained with/without the manager assumption loss $\mathcal{L}_M$.}
    \label{fig:manager assumption}
\end{figure}
\subsubsection{Expert Differentiation}
We first investigate the proposed manager assumption loss $\mathcal{L}_M$ and find that it achieves balanced training across experts, which is the premise of expert differentiation. Figure \ref{fig:manager assumption} compares the two models trained with and without $\mathcal{L}_M$, with the performance curves of different experts on the development set presented in each sub-figure. Once $\mathcal{L}_M$ is applied, four experts that fail to be trained (the performance stays around 50\% as training steps increase)  achieve comparable performance with the rest expert (expert 5 in sub-figure (a)). The result indicates that the proposed $\mathcal{L}_M$ leads balanced training across experts.
\begin{figure}[htb]
	\centering
	\includegraphics[scale=0.3]{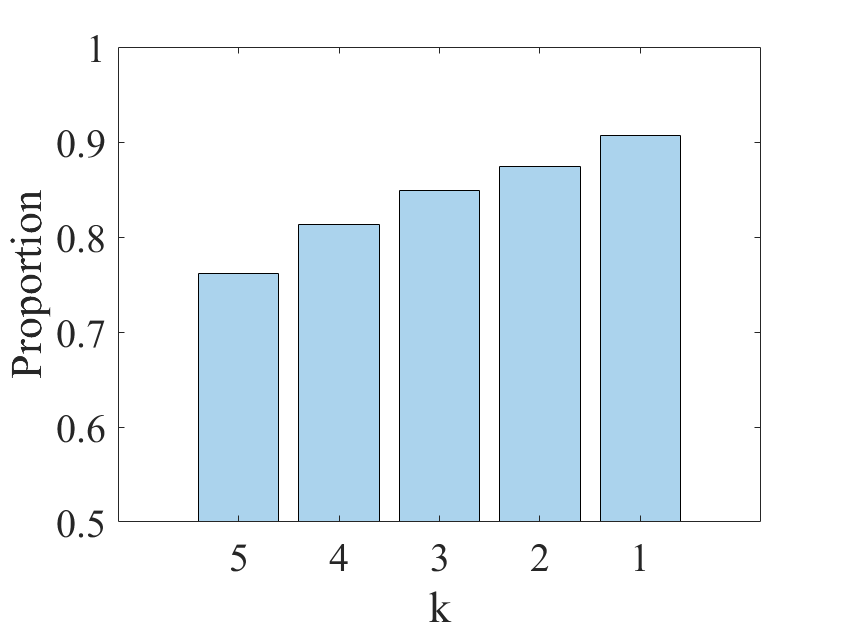}
	\caption{The proportion of statements in the test set that at least $k$ experts verify them correctly ($k\in[1,5]$).}
	\label{fig:propotion of correct}
\end{figure}

We further show that the proposed framework achieves differentiation across experts. Figure \ref{fig:propotion of correct} presents the proportion of statements in the test set that are verified correctly by at least $k$ experts ($k$ varies from 1 to 5). Note that the proportion increases rapidly as $k$ decreases (76.2\% to 90.7\% for $k$ from 5 to 1), which illustrates that experts behave differently on a large proportion of statements. The results indicate that SaMoE successfully achieves expert differentiation, which expands the original performance upper bound considerably \textbf{(90.7\%)}.
\begin{table}[htb]
\centering
\setlength{\tabcolsep}{3mm}{
	\begin{tabular}{lccc}
\hline
\multirow{2}{*}{Model} & \multicolumn{3}{c}{Accuracy} \\ \cline{2-4} 
                       & Top 1    & Top 2   & Top 3   \\ \hline
SaMoE                  & \textbf{32.0}     & \textbf{59.0}    & \textbf{76.0}    \\
SaMoE w/o Sa           & 25.4     & 44.8    & 67.6    \\ \hline
\end{tabular}}
\caption{The top-k accuracy of the management module that predicts the best-performed experts on the test set.}
	\label{table:top-k accuracy}
\end{table}
\subsubsection{Effective Attention Assignment}
We conduct a detailed analysis to investigate whether the management module assigns higher attention scores to experts with the best performance after self-adaptive learning. To achieve this goal, we regard the management module as a $n_e$-class classifier and calculate the top-k accuracy of predicting the best-performed expert (the one with the smallest cross-entropy) on the test set where k is chosen in [1, 2, 3]. The results of the analysis are presented in Table \ref{table:top-k accuracy}. The top-k accuracy is improved significantly after self-adaptive learning (\textbf{+6.6\%, +14.2\%, +8.4\%} respectively), indicating that the management module successfully assigns higher attention scores to the best-performed experts by self-adaptive learning. 

Based on the significant performance upper bound expanded by the expert differentiation, the effective attention assignment achieves more efficient cooperation across these diverse experts, thus improving the verification performance.
\section{Related Works}
\paragraph{Table-Based Fact Verification}
Most of the current models utilize programs to improve the model's ability to handle various types of numerical and logical reasoning \citep{2019TabFactA, zhong-etal-2020-logicalfactchecker, shi-etal-2020-learn,yang-etal-2020-program, yang2021exploring}, while \citet{eisenschlos-etal-2020-understanding,liu2021tapex} leverage table-based pre-trained models to parse the structural and numerical semantics of tables better. Unlike previous works, we use a novel mixture-of-experts framework to handle different logical and numerical semantics without semantic parsing and table-based pre-training.
\paragraph{Mixture of Experts}
Mixture of experts is a special model combining method. \citet{jacobs1991adaptive} first introduces this method and proposes a loss that encourages competitive learning across expert models. Afterwards, it is applied in various fields, including dialog system \citep{le2016lstm}, content recommendation\citep{ma2018modeling,zhu2020recommendation}, image classification\citep{wang2020deep,riquelme2021scaling}, etc. In this paper, we develop a self-adapted mixture-of-experts framework that achieves a more effective combination of experts by learning from the experts' performance on the train set.
\section{Conclusion}
This paper proposes a new method that exploits the mixture of experts to recognize and execute different types of reasoning required for table-based fact verification. We propose an MoE model guided with limited prior knowledge to handle different parts of the reasoning types required by table-based verification with diverse experts. Moreover, we design a supervisor network to adjust the imprecise attention score and achieve a more efficient combination across experts. A self-adaptive learning strategy is further applied to train the proposed supervisor network without prior knowledge of the task or dataset. The experiments show that the proposed model achieves a new state-of-the-art performance of 85.1\% accuracy on the benchmark dataset TABFACT. The ablation studies and analysis further indicate the effectiveness of the proposed MoE structure and self-adaptive learning strategy. We hope our work is helpful for those who aim to further exploit the power of mixture-of-experts on table-based reasoning in the future.
\section*{Acknowledgments}
We thank the anonymous reviewers for their valuable comments. This work is supported by Scientific and Technological Innovation 2030–“New Generation Artificial Intelligence” Major Project (No.2021ZD0113400), National Key Research and Development Program of China (No.2021YFC250083), and the Beijing Municipal Natural Science Foundation (No.L192026).
\bibliography{anthology,custom, custom_2}
\bibliographystyle{acl_natbib}
\clearpage
\appendix
\section{Statistics of \textsc{TabFact}} \label{apdix:statistics}
Table \ref{table:basic statistics of TABFACT dataset} shows the basic statistics of \textsc{TabFact}. As the table shows, the whole dataset is randomly divided into three subsets with the ratio be 8:1:1. The average numbers of rows and columns in tables keep approximately the same across three subsets, which reflects the consistency of data distribution.
\begin{table}[htb]
\centering
\setlength{\tabcolsep}{1mm}{
	\begin{tabular}{ccccc}
		\hline
		Split&\#Sentence&\#Table&Avg.row&Avg.col\\
		\hline
		Train    &92,283&13,182&14.1&5.5\\
		Dev    &12,792&1,696&14.0&5.4\\
		Test    &12,779&1,695&14.2&5.4 \\
		\hline
\end{tabular}}
\caption{Statistics of \textsc{TabFact}, including the number of statements, tables, and the average number of rows and columns in tables.}
	\label{table:basic statistics of TABFACT dataset}
\end{table}
\section{Parameter Initialization}\label{apdix:parameter init}
For parameter initialization, We leverage RoBERTa-Large, a pre-trained language model that has 24 transformer encoding layers. We initial parameters of the feature extractor with the embedding layer and the bottom 12 encoding layers of RoBERTa-Large and each expert with the upper 12 encoding layers of RoBERTa-Large, respectively. We use PyTorch to initialize other parameters randomly. 
\section{Specific Setting of Prior Assumption Generation}\label{apdix:specific prior}
We choose four reasoning types that appear most frequently in \textsc{TabFact}: count, comparative, superlative, and negation. The detailed definitions of four reasoning types chosen in our implementation are listed below:
\begin{enumerate}
    \item Count: counting the number of specific rows in the table, such as ``xxx be listed a total of 3 times", ``xxx win only 1 time in ...", etc.
    \item Comparative: comparing two values in the statement or cells, such as ``xxx play in more than 1 game during ...", ``xxx has a larger yyy than zzz", etc.
    \item Superlative: finding the highest/lowest value of the specific column, such as ``the longest xxx be yyy", ``the lowest score at xxx be yyy", etc.
    \item Negation: negating the original semantics of the statement, such as ``xxx has never lost a game in ...", ``xxx never score 0 points", etc.
\end{enumerate}
\begin{table}[h]
\centering
\setlength{\tabcolsep}{1mm}{
	\begin{tabular}{ccc}
		\hline
		Type&Trigger&Weight\\
		\hline
		Count    & only+[number]&1.6\\
		Count    & [number]+times&2\\
		Count    & [number]+of&1.6\\
		Count    & there be+[number]&1.6\\
		Negation    & no&1.5\\
		Negation    &not&1.5\\
		Negation    &never&1.5\\
		Negation    &didn't&1.5\\
		Comparative    & [JJS] or [RBS]&1.5\\
		Superlative    & [JJR] or [RBR]&1.5\\
		\hline
\end{tabular}}
\caption{Some trigger words/patterns applied in the generation of the prior assumption on \textsc{TabFact}.}
	\label{table:triggers}
\end{table}
A small trigger-word pool that contains only 26 trigger words/patterns is applied for the prior assumption generation: 11 triggers for the "count" type, 15 for "negation"; and for the rest types (i.e., "comparative" and "superlative" types), the NLTK package is employed to recognize the comparative and superlative words automatically. Such a small trigger-word pool injects limit prior knowledge of the dataset, indicating that the proposed method can be generalized to other datasets by simply modifying the pool of trigger words. Table \ref{table:triggers} presents some words/patterns in the trigger-word pool applied in our experiments. x+[number] denotes a combination of a word and a number that is served as a trigger (e.g., for the statement ``xxx win 3 times in ...", we match the phrase ``3 times" with the trigger ``[number]+times").
\end{document}